*Full length article*

# Longitudinal Retinal Vascular Remodeling in Myopic Children Treated with Orthokeratology or Defocus Lenses: A Two-Year Comparative Study

Zhihao Zhao [b, †], Yinzheng Zhao [b, †], Jie Zhang [a, †], Huiqin Jiang [d], Yanyu Shangguan [a], Yanfei Sun [d], Li Chen [c], Yanlong Bi [a], M.Ali Nasseri [b, e, f, *], Bing Li [a, *]

[a] Department of ophthalmology, Tongji Hospital, School of Medicine, Tongji University, Shanghai, 200065, China

[b] TUM University Hospital Rechts der Isar, Technische Universität München, 81675 Munich, Germany

[c] Department of ophthalmology, Yangpu Hospital, School of Medicine, Tongji University, Shanghai, 200090, China

[d] Department of ophthalmology, Shanghai Demu Youmei Ophthalmology Outpatient Department Co., Ltd. Shanghai, 200001, China

[e] Zhongshan Ophthalmic Center, Sun Yat-sen University, Guangzhou, 510623, China

[f] Department of Biomedical Engineering, University of Alberta, T6G 2R3 Alberta, Canada

†These authors contributed equally.

*Corresponding author(s): Prof. M.Ali Nasseri

**Abstract**

***Purposes***: To characterize longitudinal retinal vascular changes in myopic children treated with orthokeratology (OK) or multifocal defocus lenses (Defocus) and to examine their association with axial elongation.

***Methods***: In this retrospective cohort study, 43 myopic children underwent comprehensive clinical examination and fundus photography at baseline, 12 months, and 24 months. Axial length (AL) and spherical equivalent refraction (SER) were recorded at baseline, 6, 12, and 24 months. An automated segmentation model extracted vascular parameters, main vessel angle (MA), branching angle (BA), bifurcation edge angle (BEA), crossover point (COP), and terminal vessel count (TVC). Repeated-measures ANOVA assessed temporal changes. Pearson or Spearman correlations evaluated associations between AL and vascular metrics.

***Results*****:** Over 24 months, the OK group exhibited significantly slower axial elongation than the Defocus group (0.214 mm and 0.522 mm, $p < 0.01$). In the OK group, MA and BA decreased modestly, BEA in arteries declined gradually, but COP and TVC remained relatively stable. The Defocus group demonstrated more pronounced decreases in MA and BA, an increase in BEA, and significant reductions in COP and TVC ($p < 0.05$). Correlation analysis revealed stronger associations between AL and vascular parameters, especially COP and TVC, in the Defocus group at all time points, whereas only BA and BEA correlated with AL in the OK group.

***Conclusions***: OK lenses mitigate axial elongation and induce milder retinal vascular remodeling compared to Defocus lenses. Distinct temporal patterns of vascular metrics changes were observed between the two interventions, and correlate differentially with axial growth.

## 1. Introduction

Myopia has emerged as a global public health concern, with increasing prevalence among children and adolescents, particularly in East Asia.[1-3] Excessive axial elongation, referring to axial growth beyond the expected age-related physiological range and commonly accompanying progressive or high myopia, is a major risk factor for sight threatening complications,[4, 5] such as myopic maculopathy,[6] choroidal neovascularization[7] and retinal detachment.[8] Several myopia control strategies have been developed, among which Orthokeratology (OK) lenses and multifocal defocus lenses have demonstrated clinical efficacy in slowing axial elongation.[9-12] However, the underlying mechanisms of these interventions remain incompletely understood.

Recent advances in retinal imaging and image analysis have made it possible to noninvasively evaluate structural and microvascular changes in the posterior pole, including the retinal vasculature.[13-15] Retinal vascular morphology, such as vessel angles, branching complexity, and vascular density, is known to reflect underlying ocular and systemic physiological changes.[16, 17] In the context of myopia progression, alterations in vascular geometry may serve as a potential biomarker for disease severity and therapeutic response. Advanced imaging techniques, particularly optical coherence tomography angiography, have further demonstrated that myopia-related microvascular alterations may occur in different retinal and choroidal vascular layers. Collectively, these findings suggest that retinal vascular characteristics may provide complementary information beyond conventional refractive and biometric measurements. Some researchers showed that tensile force might influence changes in the retinal vascular system during axial elongation in high myopia, leading to atrophy of the vascular layer, narrowing of vessels, altered vessel angles, and reduced vascular density.[18, 19] Lim et al.[20] found that higher myopic refractive error and longer axial length were associated with a sharper bifurcation angle in arteries among middle aged and older Malaysians, along with increased branching coefficients in both arteries and veins. However, in a small cohort of pseudophakic individuals, Patton et al.[21] did not observe any association between AL and vessel bifurcation angles or connectivity indices.

Most existing studies on myopia control have focused predominantly on refractive and axial outcomes, with limited investigation into the underlying structural or vascular changes in the retina.[22, 23] Studies of myopia-control interventions have mainly focused on changes in spherical equivalent refraction and axial length, while the longitudinal patterns of retinal vascular change during treatment remain insufficiently characterized. In particular, it remains unclear whether children receiving different myopia-control interventions exhibit distinct patterns of retinal vascular remodeling over time and whether these patterns are associated with axial elongation. While retinal vascular alterations have been observed in myopic eyes, the temporal dynamics of these changes, especially in response to specific interventions, remain poorly understood. Furthermore, few studies have directly compared the impact of different myopia control strategies on the retinal microvasculature. Understanding whether these treatment modalities induce distinct patterns of vascular remodeling may offer insights into their mechanisms of action and provide potential biomarkers for treatment efficacy. This knowledge gap underscores the need for longitudinal, image-based vascular analyses to complement traditional biometric measures in myopia research.

In this study, we used an automated retinal image analysis pipeline to quantify vascular features from fundus photographs of myopic children receiving either orthokeratology or multifocal Defocus lens treatment over two years. The evaluated parameters included main vessel angle, branching angle, branching edge angle, crossover point count, and terminal vessel count. We hypothesized that the two treatment groups would exhibit distinct longitudinal patterns of retinal vascular change and that changes in these vascular parameters would be associated with axial elongation. By testing this hypothesis, the study aimed to identify potentially informative vascular parameters and provide longitudinal evidence to support the design of future prospective studies investigating retinal vascular remodeling during myopia control.

## 2. Methods

### *2.1 Study Design and Subjects*

This was a retrospective, observational cohort study conducted at the affiliated Yangpu District Central Hospital of Tongji University. We reviewed the clinical records of a consecutive series of pediatric patients who initiated either OK lens or Defocus lens treatment between January 2022 and January 2024. This consecutive sampling resulted in a total of 43 subjects who met the eligibility criteria. Treatment selection was determined before study inclusion based on clinical assessment and discussions with the children and their guardians, taking into consideration refractive status, ocular characteristics, lifestyle requirements, treatment suitability, and family preference. The study included 25 patients in the OK lens group (Eyebright Medical Technology, Beijing, Co., Ltd.) and 18 patients in the Defocus lens group (Eyepol Optical Technology, Xiamen, Co., Ltd.). Inclusion criteria were: (1) a clinical diagnosis of myopia (SER between -0.75 D and -6.00 D); (2) availability of complete clinical records and high-quality fundus photographs at baseline, 12 months, and 24 months; (3) age between 8 and 16 years at the start of treatment. Patients were excluded if they had: (1) astigmatism > 1.50 D; (2) any ocular pathology other than myopia; (3) a history of prior ophthalmic surgery; or (4) fundus images of insufficient quality for reliable automated analysis (e.g., due to poor focus, media opacity, or artifacts).

This study adhered to the principles of the Declaration of Helsinki and was approved by the Ethics Committee of Yangpu District Central Hospital (Approval Number: LL-2025-LW-003). Given the retrospective nature of the study using anonymized data, the requirement for written informed consent was waived by the committee.

### *2.2 Data Collection and Image Analysis*

Clinical data included age, sex, age at initial myopia diagnosis, SER, AL, and corneal curvature. AL and SER were recorded at baseline, 6 months, 12 months, 18 months and 24 months. Fundus photographs were acquired using Non-mydriatic Fundus Camera (Topcon, Japan) at baseline, 1 year, and 2 years. Images were captured at a 45° field of view, centered on the macula.

Retinal vascular parameters, including vessel angle, fractal dimension, vessel diameter, and vascular coefficient, were extracted from fundus images using a validated model based on segmentation and quantization. Axial elongation and refraction progression were calculated as the difference between baseline and follow-up values at each time point.

Retinal vascular parameters were extracted using a validated, automated analysis pipeline based on a deep learning segmentation model. Firstly, raw fundus images underwent preprocessing, including brightness normalization and contrast enhancement to standardize image quality. Secondly, a U-Net based deep learning architecture was employed to segment the vascular network and generate a binary vessel map. The system automatically differentiated arteries from veins. Finally, the segmented vessel map was skeletonized to extract the network's topology and then calculate the quantitative parameters. When the loss function of our data classification model converged to 0.09, the model accuracy reached 94.19%. To ensure data quality, all segmented images were visually inspected by a trained researcher, and images with significant segmentation errors were excluded. The core research focus is on comparing longitudinal trends and relative differences in vascular parameters between groups, which can reduce image magnification differences caused by axial length growth or corneal curvature changes due to OK lens use, thereby improving the accuracy of absolute measurements.

### *2.3 Statistical Analysis*

In this study, normality tests were first conducted. For continuous variables following a normal distribution, they are expressed as mean ± standard deviation (X ± S). Effect sizes were reported alongside P values where applicable. Partial eta-squared was used to describe the magnitude of effects in repeated-measures ANOVA. Correlation analyses were reported using Pearson’s correlation coefficient or Spearman’s rank correlation coefficient, together with the corresponding P values. Adjusted odds ratios and 95% confidence intervals were reported for the binary logistic regression analysis. If the data did not satisfy normality, the Friedman rank-sum test was used as an alternative. Comparisons between different treatment groups were performed using the Mann–Whitney U test. Categorical variables are presented as frequencies and percentages, with intergroup comparisons conducted using $\chi^2$ tests or Fisher's exact test. The correlation between vascular parameters and axial length, along with refractive changes, was analyzed using Pearson or Spearman correlation analysis. A binary logistic regression model was further used to evaluate the independent association between changes in vascular parameters and axial elongation, while controlling for confounding factors such as age, sex, baseline axial length, and treatment methods. The logistic regression analysis was performed to identify factors associated with greater cumulative axial elongation over the 24-month follow-up. The dependent variable was coded as 0 for cumulative axial elongation ⩽0.3 mm and 1 for cumulative axial elongation >0.3 mm. The cutoff was selected as a clinically interpretable threshold with reference to previously reported axial-growth ranges in children and its use in studies evaluating responses to myopia-control interventions. This stratification enabled the identification of demographic and retinal vascular factors associated with more pronounced cumulative axial growth. Differences were considered statistically significant when P values < 0.05. All data analyses were conducted using SPSS (version 27.0) and Python (version 3.5).

## 3. Results

A total of 43 children with myopia were included, with 25 patients in the OK lenses group and 18 patients in the Defocus lenses group (Table 1). The gender distribution was comparable between the two groups. However, patients in the OK lens group were significantly older (13.28 ± 1.81 years) than those in the Defocus lens group (11.72 ± 1.99 years, P < 0.001). Similarly, the age at first myopia diagnosis was higher in the OK lens group (10.40 ± 1.61 years) compared to the Defocus group (9.33 ± 1.64 years, P < 0.001). The OK lens group had a higher mean baseline age than the Defocus lens group. As age is closely associated with ocular growth and retinal vascular development in children, this baseline difference was considered when interpreting the longitudinal axial-length and vascular trajectories. Age was subsequently included as a covariate in the multivariable regression analysis. There were no statistically significant differences between the two groups in terms of baseline cycloplegic spherical equivalent refraction or axial length (P = 0.623; P = 0.834). Longitudinal measurements of axial length at 6, 12, and 24 months after treatment initiation also showed no significant differences between groups.

### *3.1 Changes in Axial Length between Different Groups*

The longitudinal changes in AL between two groups were evaluated at multiple follow-up time points. As shown in Figure 2A, both groups exhibited progressive axial elongation over the 24-month period. However, the rate of axial length increase was significantly lower in the OK lens group compared to the Defocus lens group. At 6, 12, and 24 months, the mean AL increases from baseline in the OK group were 0.093 mm, 0.132 mm, and 0.214 mm, respectively, whereas the corresponding increases in the Defocus group were 0.066 mm, 0.203 mm, and 0.522 mm.

Additionally, to better understand the short-term dynamics of axial elongation, inter-visit AL differences were plotted (Figure 2B). Between 6 and 12-month follow-ups, the Defocus group showed a continued and relatively stable increase in AL (0.138 mm), while the OK lens group exhibited a lower increment (0.039 mm). During the second year (months 12 to 24), the Defocus lens group again showed a notable increase in AL (0.204 mm), in contrast to the OK group, which maintained minimal elongation (0.089 mm).

### *3.2 Longitudinal Changes in Major Vascular Parameters Across Different Groups*

Repeated-measures ANOVA revealed statistically significant interaction effects between group and time for main vascular parameters, indicating distinct longitudinal trends between the OK lens and Defocus lens groups ($P < 0.001$).

The temporal trends of vascular parameters were analyzed separately for arterioles and venules in both treatment groups (Figure 3). In the OK lens group, the arterial MA showed a slight and relatively stable decrease over the 24-month follow-up period (Figure 3A). In contrast, the Defocus lens group exhibited a more pronounced reduction in arterial MA, particularly after the first year. Venular MA remained relatively stable in both groups.

The branching angle (BA) of retinal vessels demonstrated a decreasing trend in both treatment groups (Figure 3B). In the OK lens group, both arteriolar and venular BA values showed a steady and progressive decline throughout the observation period. In the Defocus lens group, arteriolar BA also decreased over time, with a mild reversal trend between 12 and 24 months. Interestingly, venular BA remained relatively stable, with a slight decrease observed in the second year.

The bifurcation edge angle (BEA) exhibited distinct temporal patterns across groups and vessel types (Figure 3C). In the OK lens group, venular BEA remained relatively stable, while arteriolar BEA showed a progressive and marked decrease throughout the follow-up. In contrast, the Defocus lens group demonstrated an obvious decreasing trend in arteriolar and increasing trend in venular BEA values, particularly between 12 and 24 months.

As shown in Figure 3D, the mean number of crossover points (COP) exhibited distinct temporal patterns between the OK lens and Defocus lens groups. In both arterial and venous measurements, the previous group showed relatively stable or mildly increasing values over the 24-month follow-up period. The second group exhibited an initial decline in both arterial and venous crossover point counts at 12 months, which then stabilized through 24 months.

Figure 3E demonstrates the longitudinal changes in terminal vascular counts (TVC) across both treatment groups. At all measured time points, a gradual reduction in terminal vessel numbers was observed in both groups. However, the decline was more pronounced in the Defocus lens group, particularly at 12 months, with minimal recovery observed thereafter.

### *3.3 Correlation Between Axial Length and Vascular Parameters*

Correlation analyses between AL and retinal vascular parameters at each follow-up time point revealed group-specific patterns (Table 2). In the Defocus lens group, AL was significantly correlated with most arterial and venous parameters across all time points, particularly with COP and TVC (P <

0.05). Significant correlations were also observed between AL and MA at all time points ($P < 0.05$), while other parameters such as BA, BEA, and BEC showed inconsistent associations.

The OK lens group exhibited fewer significant correlations (Table 3). For arterial parameters, only the BA and BEA consistently showed correlations with AL ($P < 0.05$), while COP and TVC were only partially correlated at baseline and early follow-up. For venous parameters, correlations with AL were generally weak and less consistent in the OK lens group.

A binary logistic regression analysis was performed to evaluate the independent association between changes in retinal vascular parameters and axial elongation, controlling for confounding factors such as age, gender, and baseline axial length (Table 4). The dependent variable was coded as 0 for AL changes ≤0.3 mm and 1 for changes >0.3 mm. This cutoff was used as a pragmatic analytical threshold and was not considered a universally established clinical definition of rapid axial elongation. Age itself is a key factor influencing the progression of myopia. To correct for this confounding effect, we adjusted for age as a covariate in the regression analysis. Independent variables included age, gender, and various morphological parameters for both arteries and veins. The regression analysis revealed that arterial BA were statistically significant predictors ($P < 0.05$). After controlling for all other variables, the odds of significant myopia progression increased by 2.3 times for every one-year increment in age. The OR value less than 1 for arterial BA indicates a negative correlation between arterial BA and significant axial elongation, suggesting that a larger arterial BA may have a protective effect against excessive axial elongation.

Collectively, these findings demonstrate distinct and quantifiable longitudinal patterns of retinal vascular remodeling across the two treatment groups. Although the clinical significance and applicable thresholds of these vascular changes require further validation, the identified parameters provide an empirical basis for subsequent studies evaluating their potential value in monitoring axial growth and myopia-control outcomes.

## 4. Discussion

Retinal fundus imaging offers a noninvasive and reproducible approach combined with quantitative vascular analysis to assess microvascular alterations in vivo. The retinal vasculature shares embryological, anatomical, and physiological characteristics with the cerebral and systemic microcirculation, making it an accessible surrogate for evaluating vascular health.[24, 25] Quantitative analysis of retinal vascular features, including vessel diameter,[26] fractal dimension,[27] and vascular density, has been increasingly utilized to detect early vascular changes in a variety of ocular and systemic diseases, such as diabetic retinopathy, hypertensive retinopathy, glaucoma, and age-related macular degeneration.[28-30]

Our findings reveal distinct patterns of retinal vascular remodeling associated with different myopia control strategies. These vascular changes may reflect underlying differences in the biomechanical and physiological responses to axial elongation modulated by optical interventions. By quantifying these changes at multiple time points, fundus-based vascular analysis not only enables the identification of early biomarkers of disease progression but also allows for monitoring treatment effects and comparing different therapeutic modalities.

Compared with the defocus lens group, the OK lens group exhibited significantly less axial elongation over 24 months. This finding is broadly consistent with previous studies reporting slower axial elongation among children treated with OK lenses.[10] Notably, vascular structural changes, including reductions in BA, TVC and COP, were generally milder in the OK lens group, suggesting that axial elongation may play a mechanistic role in the stability of peripheral retinal vessels and their nearly normal branching morphology.[31] The reduced mechanical stretch on the retina and sclera likely minimizes tractional forces on retinal vessels, thus preserving their geometric configurations. Furthermore, the relative stability of vascular parameters may reflect preserved retinal perfusion and autoregulatory function under slower myopic progression. The coordinated longitudinal patterns observed in axial length and retinal vascular morphology indicate a close association between ocular growth and vascular remodeling during myopia-control treatment. Changes in retinal geometry, tissue tension, perfusion demand, and vascular autoregulation may provide biologically plausible contexts for understanding these longitudinal associations. The decline in vascular COP and TVC signifies peripheral capillary rarefaction, a reduction that was notably more prominent in the Defocus lens group. This pattern is frequently associated with progressive myopia, likely attributed to mechanisms such as tissue remodeling, diminished pro-angiogenic cues, or localized hypoxic conditions resulting from retinal stretching or compromised metabolic support in the periphery. The diminished topological complexity of the retinal vascular network may affect local perfusion and oxygenation.

Venular BA in the Defocus group showed an obvious decrease, but arterial BA increased, which may represent compensatory dilation or vascular adaptation to maintain retinal perfusion in the setting of progressive structural remodeling. However, the OK group showed more gradual, controlled changes, consistent with attenuated retinal distortion. This may reflect more localized and controlled remodeling in peripheral vascular system under OK lens treatment. A study involving 493 patients indicated that the higher the degree of myopia, the smaller the artery branching angle. Different refractive errors in myopia are associated with different vascular bifurcation patterns. Conversely, the Defocus lens group exhibited increased BEA in venous branches and decreased BEA in arterial branches, suggesting peripheral vessel dilation or disruption of normal branching geometry. The OK lens group showed smaller changes. This divergence may reflect differing impacts of the two treatments on peripheral retinal biomechanics or oxygen demand. Li et al.[32] also indicates a reduced density of retinal micro vessels in both the superficial and deep vascular plexuses among individuals with high myopia. This conclusion aligns with our findings.

Despite the valuable insights provided by our findings, several limitations of this study should be acknowledged. First, the study failed to control for other potential confounding factors, such as circadian variations in axial length, systemic vascular health, or lifestyle habits, all of which could influence retinal vascular morphology. Second, the sample size was small, and the retrospective design analyzing consecutive cases within a specific time period limits the generalizability of the findings to a broader population and prevents establishing causality. Although our analysis provides valuable preliminary longitudinal evidence regarding retinal vascular development in myopic children undergoing treatment, future prospective studies incorporating age-matched untreated, non-myopic, and alternative treatment control groups are warranted. Our results should be regarded as exploratory associations rather than definitive treatment effects. Given these limitations, future research directions are clear. We will design large scale, age matched prospective randomized controlled trials combined with more advanced imaging techniques such as optical coherence tomography angiography (OCT-A) to obtain quantitative information on retinal vascular networks at different layers, thereby deepening our understanding of vascular remodeling mechanisms.

## 5. Conclusion

In summary, this study provides preliminary evidence indicating that OK lens treatment is associated with slower axial elongation and less pronounced retinal vascular remodeling over a two-year follow-up period compared to defocus lens treatment. These observed associations lay a foundation for the design of larger, more rigorous prospective clinical trials and help identify vascular parameters that merit further investigation. Future large scale, age-matched prospective studies utilizing proper control groups and multimodal imaging modalities are required to validate these findings and elucidate their potential clinical implications.

## Abbreviations

OK Orthokeratology
AL Axial length
SER Spherical Equivalent Refraction
MA Main vessel angle
BA Branching angle
BEA Bifurcation edge angle
BEC Bifurcation edge coefficient
COP Crossover point
TVC Terminal vessel count

## Tables

**Table 1.** Clinical Characteristics and Baseline Parameters

| Characteristics | Groups | | P Value |
|---|---|---|---|
| | OK Lens (n=25) | Defocus Lens (n=18) | |
| Gender | | | 0.435 |
| Male | 16 | 10 | |
| Female | 9 | 8 | |
| Age | 13.28±1.81 | 11.72±1.99 | < 0.001 |
| First Diagnosis Age | 10.40±1.61 | 9.33±1.64 | < 0.001 |
| Baseline Values | | | |
| Cycloplegic Refraction | -2.45±1.08 | -2.69±2.48 | 0.623 |
| Axial Length | 24.44±0.85 | 24.39±0.85 | 0.834 |
| Longitudinal Values | | | |
| Axial Length After 6 months | 24.54±0.81 | 24.46±0.92 | 0.872 |
| Axial Length After 12 months | 24.58±0.80 | 24.60±0.90 | 0.606 |
| Axial Length After 24 months | 24.75±0.99 | 24.92±0.64 | 0.595 |

**Table 2.** Analysis of Correlation Between Axial Length and Retinal Arterial Vascular Parameters at Different Time Points

| Artery | MA | BA | BEA | BEC | COP | TVC |
|---|---|---|---|---|---|---|
| Defocus Lens Group | | | | | | |
| T0 Axial Length | < 0.05 | 0.998 | 0.626 | 0.673 | < 0.05 | < 0.05 |
| T6m Axial Length | < 0.05 | 0.961 | 0.228 | 0.939 | < 0.05 | < 0.05 |
| T12m Axial Length | < 0.05 | 0.901 | 0.474 | 0.665 | < 0.05 | < 0.05 |
| T24m Axial Length | < 0.05 | 0.499 | 0.173 | 0.860 | < 0.05 | < 0.05 |
| OK Lens Group | | | | | | |
| T0 Axial Length | 0.783 | < 0.05 | < 0.05 | 0.158 | 0.556 | 0.616 |
| T6m Axial Length | 0.881 | < 0.05 | < 0.05 | 0.160 | 0.494 | 0.555 |
| T12m Axial Length | 0.880 | < 0.05 | < 0.05 | 0.162 | 0.605 | 0.691 |
| T24m Axial Length | 0.738 | < 0.05 | < 0.05 | 0.154 | 0.258 | 0.324 |

*MA: main angle; BA: branching angle; BEA: bifurcation edge angle; BEC: Bifurcation edge coefficient; COP: crossover point; TVC: terminal vessel count*

**Table 3.** Analysis of Correlation Between Axial Length and Retinal Venous Vascular Parameters at Different Time Points

| Vein | MA | BA | BA (asymmetry) | BEA | BEA (asymmetry) | COP | TVC |
|---|---|---|---|---|---|---|---|
| Defocus Lens Group | | | | | | | |
| T0 Axial Length | 0.360 | 0.894 | 0.561 | 0.157 | < 0.05 | < 0.05 | < 0.05 |
| T6m Axial Length | 0.574 | 0.856 | 0.623 | 0.095 | 0.094 | < 0.05 | < 0.05 |
| T12m Axial Length | 0.648 | < 0.05 | < 0.05 | 0.154 | < 0.05 | < 0.05 | < 0.05 |
| T24m Axial Length | 0.098 | < 0.05 | < 0.05 | 0.148 | < 0.05 | < 0.05 | < 0.05 |
| OK Lens Group | | | | | | | |
| T0 Axial Length | 0.440 | 0.731 | 0.680 | 0.478 | 0.698 | 0.758 | 0.722 |
| T6m Axial Length | 0.335 | 0.685 | 0.744 | 0.066 | 0.286 | 0.391 | 0.377 |
| T12m Axial Length | 0.315 | 0.627 | 0.621 | 0.515 | 0.767 | 0.733 | 0.719 |
| T24m Axial Length | 0.089 | 0.665 | 0.861 | 0.483 | 0.791 | 0.474 | 0.449 |

*MA: main angle; BA: branching angle; BEA: bifurcation edge angle; COP: crossover point; TVC: terminal vessel count*

**Table 4.** Binary Logistic Regression Analysis for Factors Associated with Significant Axial Length Change

| | B | S.E. | Wald | Sig. (P value) | Exp (B) (OR) | Exp (B) 95% CI | |
|---|---|---|---|---|---|---|---|
| | | | | | | Lower Limits | Upper Limits |
| Age | 0.842 | 0.298 | 8.007 | 0.005 | 2.322 | 1.296 | 4.162 |
| Gender | -0.496 | 0.891 | 0.31 | 0.578 | 0.609 | 0.106 | 3.494 |
| AL Baseline | -0.297 | 0.552 | 0.288 | 0.591 | 0.743 | 0.252 | 2.195 |
| Artery MA | -0.019 | 0.015 | 1.588 | 0.208 | 0.982 | 0.954 | 1.01 |
| Artery BA | -0.084 | 0.041 | 4.137 | < 0.05 | 0.919 | 0.848 | 0.997 |
| Artery BEA | 0.021 | 0.014 | 2.137 | 0.144 | 1.021 | 0.993 | 1.05 |
| Artery COP | -0.352 | 0.401 | 0.772 | 0.38 | 0.703 | 0.321 | 1.542 |
| Artery TVC | 0.236 | 0.382 | 0.382 | 0.537 | 1.266 | 0.599 | 2.677 |
| Vein MA | 0.012 | 0.016 | 0.551 | 0.458 | 1.012 | 0.981 | 1.044 |
| Vein BA | -0.047 | 0.036 | 1.708 | 0.191 | 0.954 | 0.89 | 1.024 |
| Vein BEA | -0.021 | 0.018 | 1.327 | 0.249 | 0.98 | 0.946 | 1.015 |
| Vein COP | -0.462 | 0.285 | 2.64 | 0.104 | 0.63 | 0.361 | 1.1 |
| Vein TVC | 0.56 | 0.293 | 3.651 | 0.056 | 1.751 | 0.986 | 3.110 |

*95% CI: 95% Confidence Interval；S.E.: Standard Error；P<0.05 means significant*

## Figure Captions

**Figure 1.** Flowchart of the Study Design and Data Analysis Pipeline. This study included 43 subjects with longitudinal images after applying exclusion criteria, divided into the OK lens group and Defocus lens group based on their treatments. Arteries and veins were extracted from the corresponding images using a segmentation model, and retinal vessel parameters were obtained through an automated quantification system. These parameters were then analyzed for correlation with changes in axial length of the eye.

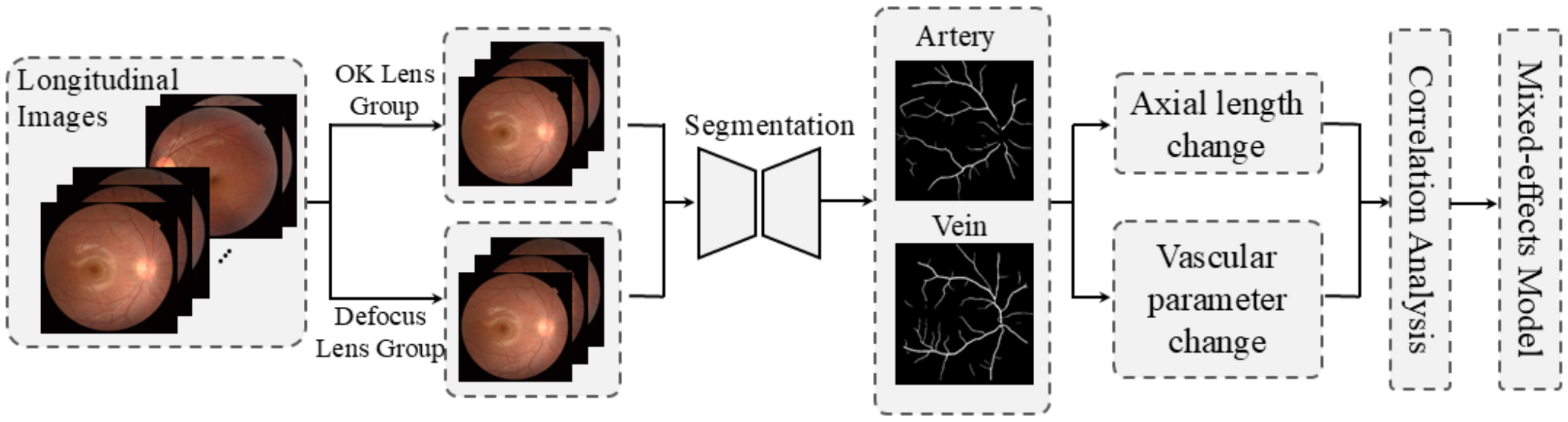

**Figure 2.** Longitudinal changes in axial length over 24 months in the OK lens group and Defocus lens group. (A) shows changes in axial length compared to baseline, and (B) displays changes in axial length between adjacent follow-up time points. The red line represents the OK lens group, and the blue line represents the Defocus lens group.

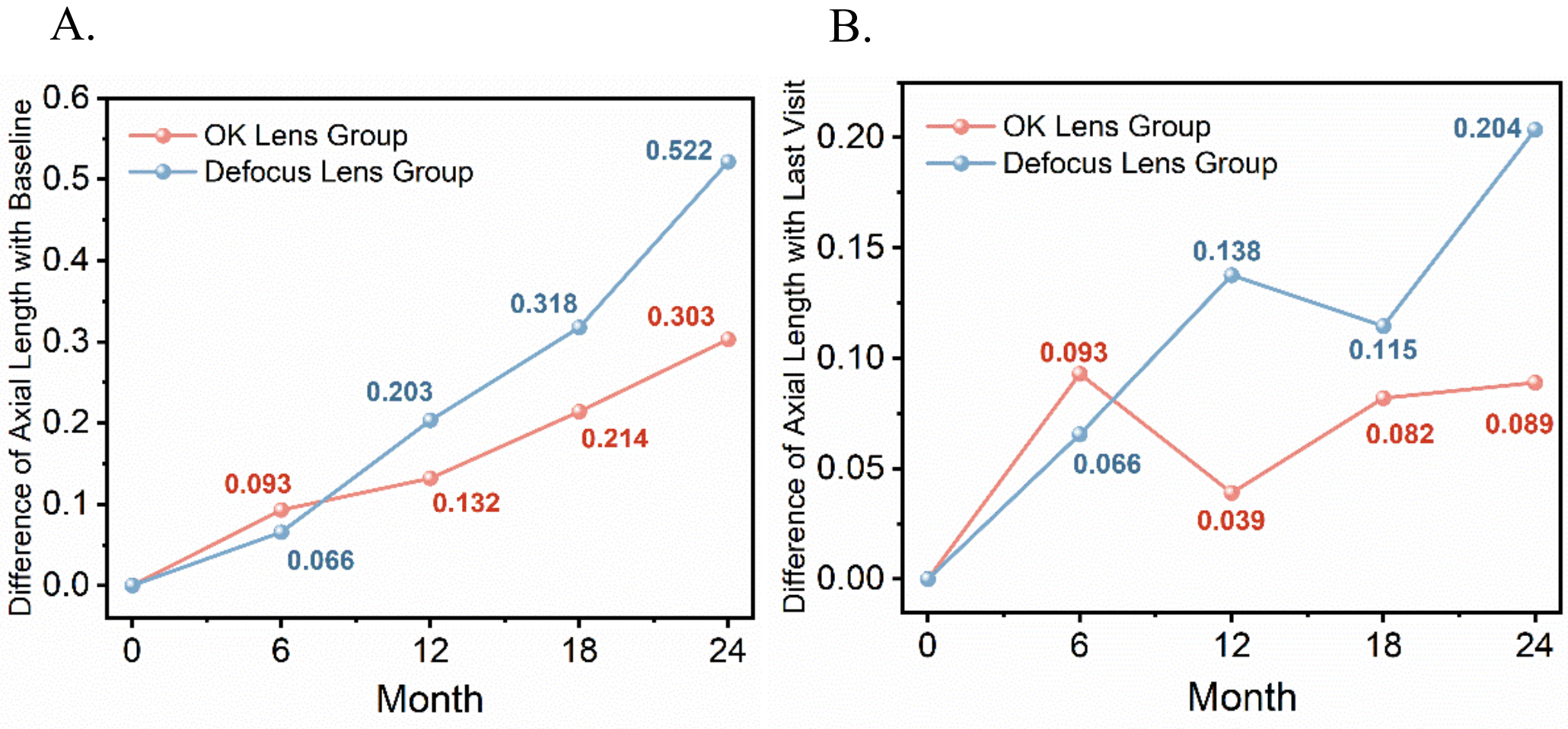

**Figure 3.** Changes in retinal artery and vein parameters over time across different groups. Red lines represent arteries; blue lines represent veins. Solid lines indicate the OK lens group, while dashed lines indicate the Defocus lens group. The analysis of retinal vessel parameters primarily focused on five aspects: (A) main vessel angle, (B) branching angle, (C) bifurcation edge angle, (D) number of crossover points, and (E) number of terminal vessels, expressed as averages.

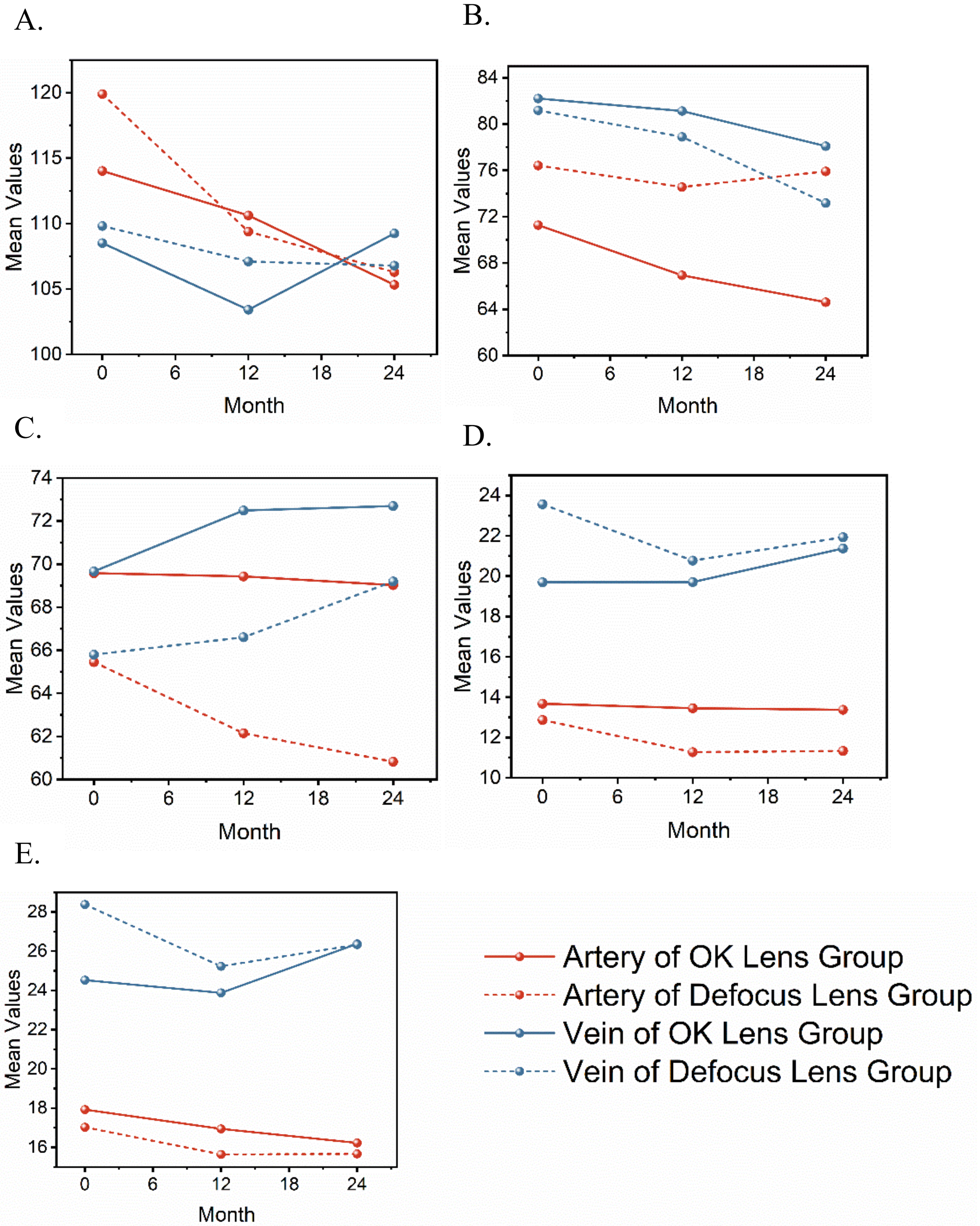